\documentclass[11pt,a4paper]{article}

\usepackage[T1]{fontenc}
\usepackage[utf8]{inputenc}
\usepackage{lmodern}
\usepackage[margin=2.5cm]{geometry}
\usepackage{microtype}
\usepackage{graphicx} 
\usepackage{float} 
\usepackage{caption} 
\usepackage{subcaption} 
\usepackage{verbatim} 
\usepackage{listings} 
\usepackage[square,sort,comma,numbers]{natbib} 
\usepackage{amsmath} 
\usepackage{authblk}
\usepackage{eso-pic}
\usepackage[hidelinks]{hyperref}
\newcommand{\WorldScientificAcceptedManuscriptNotice}{%
  \AddToShipoutPictureFG*{%
    \AtPageLowerLeft{%
      \put(44,10){%
        \parbox[b]{507pt}{%
          \hrule
          \vspace{2pt}
          \fontsize{6.0}{6.8}\selectfont
          \raggedright
          Electronic version of an article published as
          \textit{International Journal of Semantic Computing}, Vol.~11,
          No.~4 (2017), pp.~473--496,
          \href{https://doi.org/10.1142/S1793351X17400190}{%
          \textcolor{blue}{https://doi.org/10.1142/S1793351X17400190}}
          \textcopyright~World Scientific Publishing Company,
          \href{https://www.worldscientific.com/worldscinet/IJSC}{%
          \textcolor{blue}{https://www.worldscientific.com/worldscinet/IJSC}}.
        }%
      }%
    }%
  }%
}

\begin{document}
\WorldScientificAcceptedManuscriptNotice

\title{REASONING-SUPPORTED ROBUSTNESS VALIDATION\\OF AUTOMOTIVE E/E COMPONENTS}

\author[1]{Jan Novacek}
\author[1]{Alexander Viehl}
\author[2]{Oliver Bringmann}
\author[2]{Wolfgang Rosenstiel}
\affil[1]{FZI Forschungszentrum Informatik,\\
Haid-und-Neu-Str. 10-14, 76131 Karlsruhe, Germany}
\affil[2]{Eberhard Karls Universit\"at T\"ubingen,\\
Sand 14, 72076 T\"ubingen, Germany}
\date{}

\maketitle

\begin{center}
\small
\textsuperscript{1}\href{mailto:novacek@fzi.de}{\texttt{novacek@fzi.de}}\quad
\href{mailto:viehl@fzi.de}{\texttt{viehl@fzi.de}}\\[2pt]
\textsuperscript{2}\href{mailto:bringman@informatik.uni-tuebingen.de}{\texttt{bringman@informatik.uni-tuebingen.de}}\quad
\href{mailto:rosenstiel@informatik.uni-tuebingen.de}{\texttt{rosenstiel@informatik.uni-tuebingen.de}}
\end{center}

\begin{center}
\small Received 14 April 2017; revised 12 June 2017; accepted 1 June 2017.
\end{center}

\begin{abstract}
This article presents an ontology-supported approach to tackle the complexity of the Robustness Validation (RV) process of automotive electrical/electronic (E/E) components. The approach uses formalized knowledge from the RV process and stress, operating, and load profiles, so-called Mission Profiles (MPs). In contrast to the error-prone industrially established manual procedure, we show how component characteristics are formalized in OWL in order to form the foundation of an efficient automated analysis selection and decision support during the RV process. Additionally, a rule-based transformation of component characteristics upon propagation via SWRL is described. The proposed approach is based on the idea of mapping MPs to an OWL representation in order to allow to execute semantic queries against MP data to improve their integration into the RV process. The resulting ontology-supported application framework has been applied to an industrial use-case from automotive power electronics. A generalization of the approach is described and demonstrated by applying it to stress test selection within the AEC Q100 standard. We present experimental results showing that the RV process can be significantly improved in terms of reduced design time and increased exhaustiveness by automating the analyses selection step and the provisioning of all the relevant data to be used.
\end{abstract}


\section{Introduction}
\label{sec:introduction}
Highly automated driving functions are revolutionizing the automotive domain by promising unique functional and efficiency selling points as well as to improve safety, as 97 \% of all accidents are due to human errors \cite{hoeger2008highly}.
Improved safety is - besides comfort and efficiency - the reason for Advanced Driver Assistance Systems (ADAS) being a well accepted option in premium vehicles \cite{aeberhard2015experience} but is also becoming common for other market segments. ADAS also constitute the foundation for highly-automated driving functions.
Thus, automotive Original Equipment Manufacturers (OEMs) face increasing time and cost pressure while at the same time complexity is tremendously increasing. Moreover, environmental conditions for the electrical/electronic (E/E) components are harsh. Consequently, quality, functionality and reliability of automobiles have become the decisive factors of global competitiveness in the automotive industry today \cite{byrne2008handbook}.

\emph{Robustness Validation} (RV) is a process that demonstrates that a product performs its intended functions with sufficient margin under certain specified environmental conditions \cite{byrne2008handbook}. RV involves the creation of a \emph{robustness validation plan}. Within this step, a developer needs to identify all necessary analyses for a component which are required to prove robustness. RV is a means to improve reliability which is why it is gaining increasing importance.

The environment of an E/E component is fundamental for its lifetime and operation. At temperatures above 100 $^{\circ}$C, too high current density can cause electromigration for example \cite{black1967mass}. Environmental conditions can be captured in so-called \emph{Mission Profiles} (MPs). They play an important role during the design and development of automotive systems, especially with regard to the propagation of requirements along the process chain of development \cite{nirmaier2014mission,abelein2012complexity}. MPs are simplified representations of relevant conditions to which a component will be exposed to throughout its whole life cycle \cite{byrne2008handbook}. They contain various relevant stress factors and have proven their applicability in cross-domain constraint transformation \cite{katzschke2014application}. MPs provide data required for certain analyses in the RV process.

The creation of the robustness validation plan is usually high error-prone due to the fact that requirements for analyses may be very complex or even change over time. Hence, necessary analyses might be easily missed out or an analysis could be conducted which is not required. RV and MP consideration is still mainly a manual task, today \cite{jerke2014mission}.
Another problem is the fact that it is hard to consolidate models, data and analyses methodologies of non-functional properties.
Prior to de-facto standardization of the RV process, there did not exist any specification of conditions for RV analyses. That is why automating this process was not reasonable.
While there exists a format for MPs, it currently provides only little semantic information. Its purpose is primarily to serve as a container for data. Conducting an analysis therefore requires searching the data manually for relevant portions.

Creating effective Knowledge Management Systems is a key success factor in engineering process improvement \cite{chourabi2008ontologySE}. This work is concerned with the automation of a step in the RV processes to decrease required manual effort by automatically selecting analyses and providing the corresponding MP data. Part of this automation is the consideration of the component characteristics propagation across component mounting points. While we describe an approach to automating the RV process utilizing ontologies and reasoning, main contributions of this work are:
\begin{itemize}
\item Approach to decreasing complexity of the RV process.
\item Mapping MPs to OWL representations.
\item Formalization of domain knowledge in OWL.
\item Propagation of component characteristics.
\end{itemize}

This paper is structured as follows: The RV process and the characteristics of the industrially established RV process flow w.r.t. this work is further described in section \ref{sec:background} together with more information on MPs. Section \ref{sec:related_work} describes related work in the fields of Systems Engineering and Mission Profile Aware Design. The main approach is then described in section \ref{sec:approach}. Experimental results are presented in section \ref{sec:example_application} and section \ref{sec:generalization} describes the generalization of the presented approach. Finally, section \ref{sec:conclusions} includes next steps, limitations and conclusions.

\section{Background}
\label{sec:background}
In this section, a short introduction to Robustness Validation and Mission Profiles is given with regard to this work.

\subsection{Robustness Validation}
\label{sec:background_robustness_validation}
\begin{figure}
  \centering
  \includegraphics[width=0.45\linewidth]{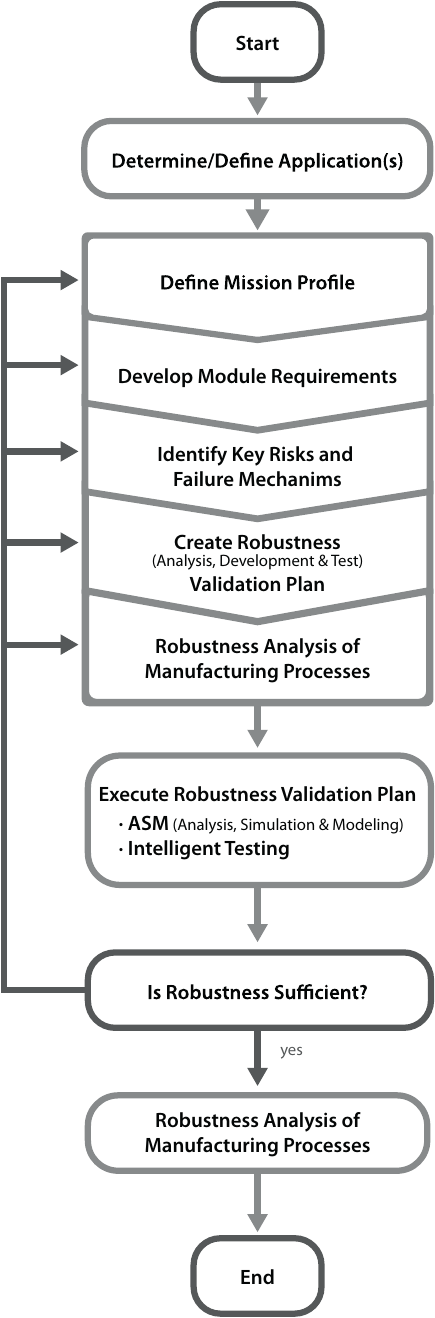}
  \caption[bla]{Robustness Validation process flow, adapted from ~\cite{byrne2008handbook}}
  \label{fig:robustness_validation_process_flow}
\end{figure}
When doing Robustness Validation (RV) developers need to conduct various analyses with their Device Under Test (DUT), for example a \emph{Physical Stress Analysis}. Right now, these analyses are done manually for the most part and the available MPs also need to be searched manually for the required data. Moreover, a developer needs to select required analyses manually.

The RV process flow as defined in the Handbook for Robustness Validation of Automotive Electrical/Electronic Modules \cite{byrne2008handbook} consists of various steps, from the determination/definition of the application(s) to production monitoring, see Fig. \ref{fig:robustness_validation_process_flow}.

A step in the RV process flow is to identify analyses which need to be done for a certain component or system, namely the creation of the \emph{robustness validation plan} (step 6 on the system level \cite{byrne2014handbookSystemLevel}). In this step a developer needs to identify all required analyses which are needed to approve robustness according to the Handbook for Robustness Validation. As stated before, this step involves manual action of the developer.
To measure the complexity of the RV process a metric such as the NOA (number of activities, see for example \cite{cardoso2006discourse}) in this process could be used.

\subsection{Mission Profiles}
\label{sec:mission_profiles}
\begin{figure}
  \centering
  \includegraphics[width=.8\linewidth]{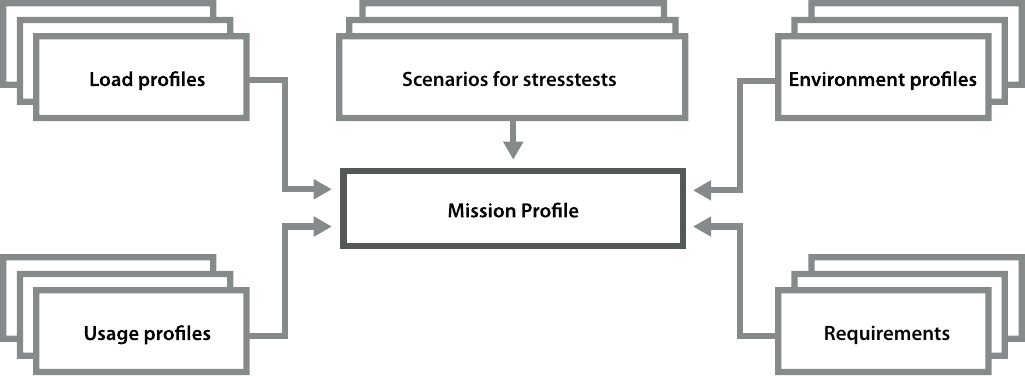}
  \caption{Mission Profile composition}
  \label{fig:mission_profile_composition}
\end{figure}

Mission Profiles (MPs) "define the application specific context for a certain component" \cite{nirmaier2014mission}. They are simplified representations of relevant conditions to which a component will be exposed to throughout its whole life cycle \cite{byrne2008handbook}. As mentioned in the introduction, MPs contain various relevant stress factors - they are composed of different sources, see Fig. \ref{fig:mission_profile_composition}.
There exists a format for MPs, which has been developed in the ResCar 2.0 project\footnote{https://www.edacentrum.de/rescar/}. It consists of three layers (core, template and extensions) to implement modularity which is further described in \cite{nirmaier2014mission}. Although the format is not yet an official standard, the autoSWIFT project\footnote{https://www.edacentrum.de/autoswift/} aims to further improve the format and standardize it.

Depending on its type, a MP may be very large and the information stored in it can be diverse, containing temperature data, vibration data and so on. In case an analysis in the RV process needs to be done, the analyst needs to search for the required data. That is also why we think using semantic technologies would be beneficial, as they would enable a better integration of this data in the RV process.

\subsubsection{Integration of the PMML}
\begin{figure}
  \centering
  \begin{subfigure}{.465\textwidth}
 	\includegraphics[width=\linewidth]{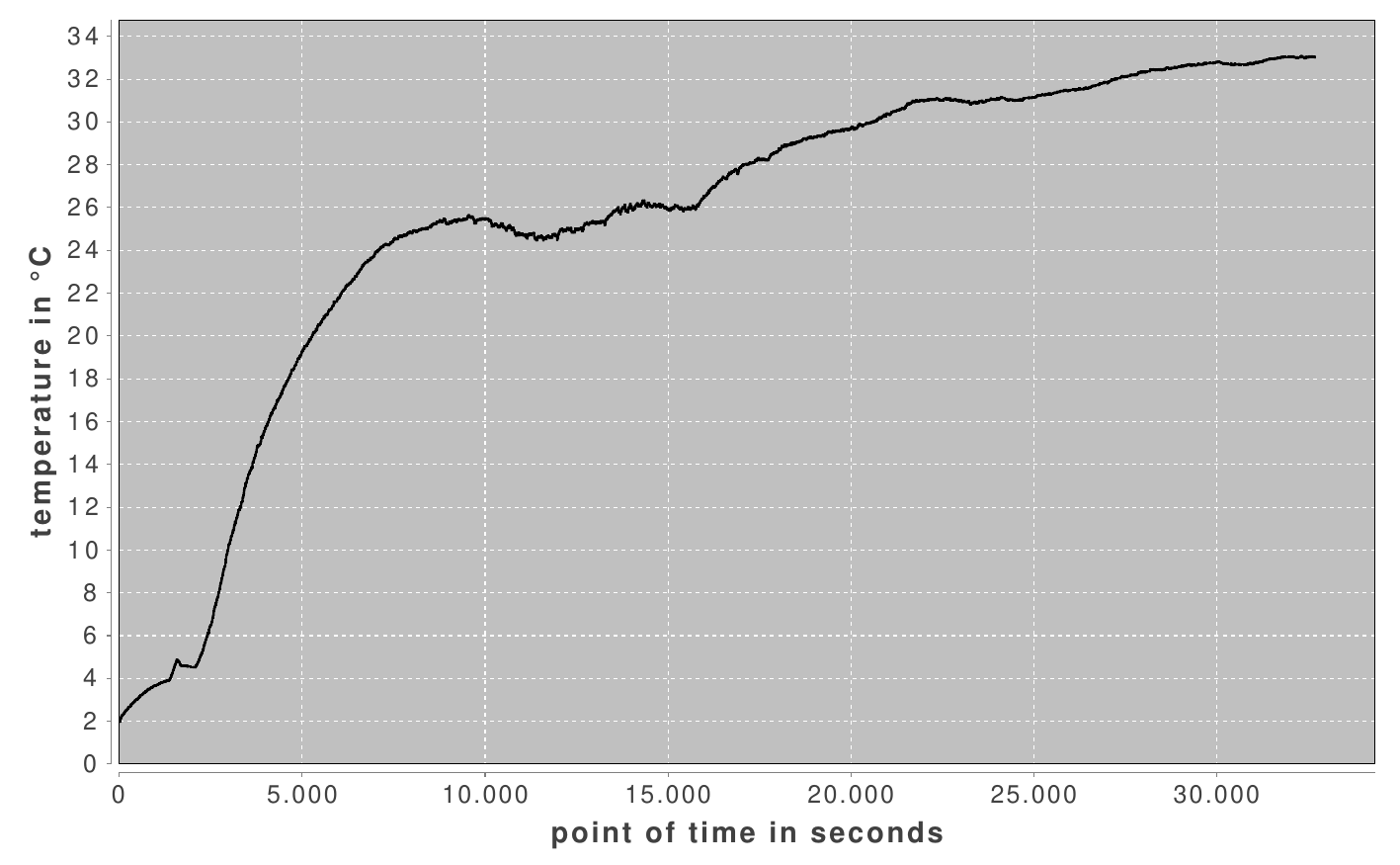}
 	\caption{Temperature profile}
  	\label{fig:temperatureA}
  \end{subfigure}
  \begin{subfigure}{.465\textwidth}
 	\includegraphics[width=\linewidth]{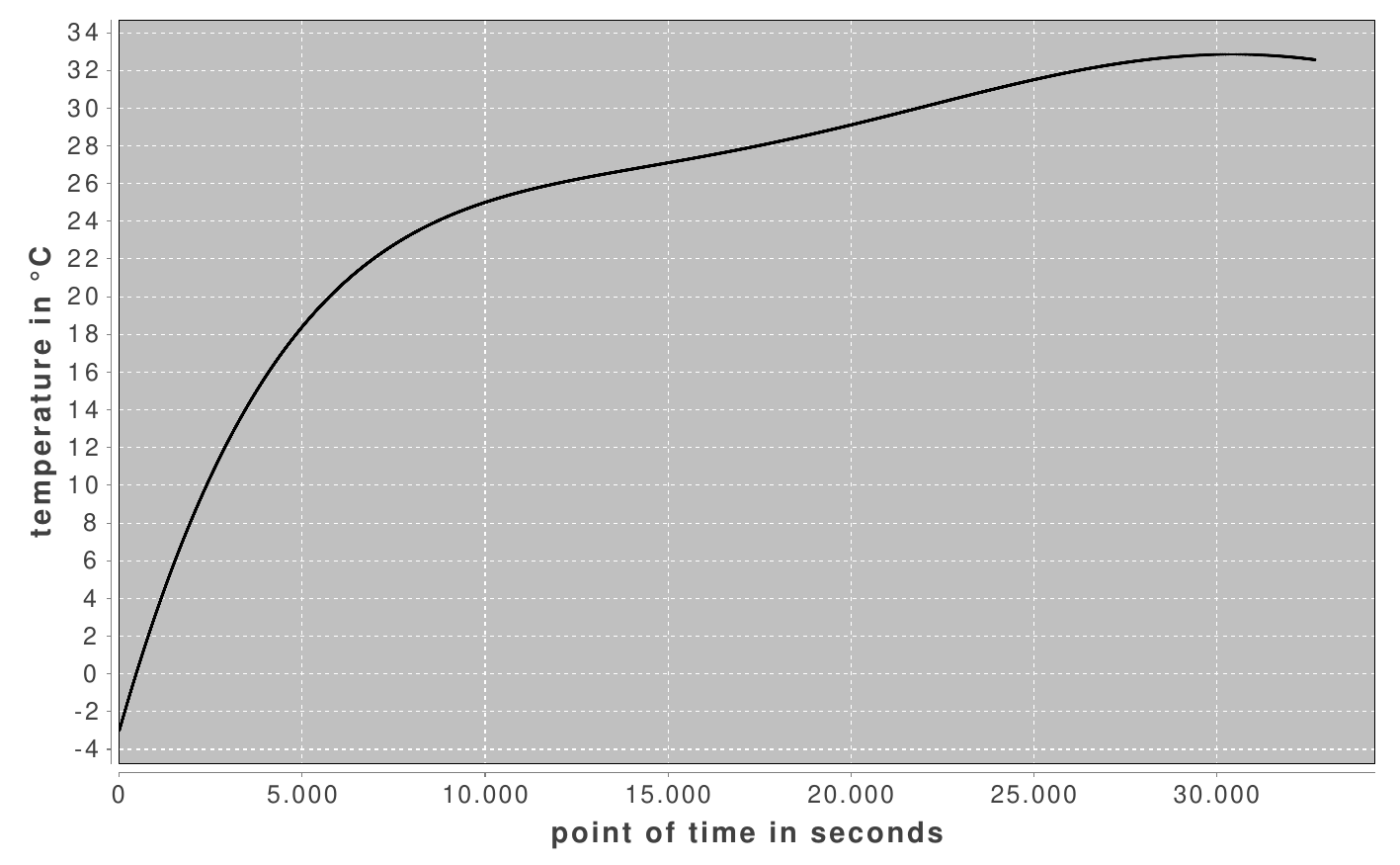}
 	\caption{Calculated model}
  	\label{fig:temperatureB}
  \end{subfigure}
  \caption{Temperature profile and calculated regression model}
  \label{fig:temperatures}
\end{figure}

The Predictive Model Markup Language (PMML) \cite{guazzelli2009pmml} is currently being integrated into the MP format primarily for the transportation of meta-models. One advantage of this integration is the replacement of enormous amounts of data by providing a model for the computation of values instead. So for example instead of listing all sampled values over time of a temperature profile, a model could be used which describes this functional load as a function. In opposition to just querying the MP for the required data the function is then used to calculate the data on the fly. Fig. \ref{fig:temperatures} shows an example of a plotted calculated regression model in comparison to the measurement data. The calculated model is represented by a polynomial function of fourth degree: $f(x)=c_0+c_1x+c_2x^2+c_3x^3+c_4x^4+c_5x^5$ where the regression coefficients $c_i, i \in \{0, ..., 5\}$ have been fit using the method of least squares. PMML allows the transportation of this model and the integration into the MP format the specification of a functional load as such a function.

On the technical level, the integration of the PMML into the MP format is realized by specifying the \emph{xsd:any}\footnote{https://www.w3.org/TR/xmlschema-0/\#any} element in the MP XSD. This element allows the extension of the format with elements not specified by the MP format schema.

\section{Related Work}
\label{sec:related_work}

\subsection{Systems Engineering}
Van Ruijven presented an ontology for Systems Engineering in \cite{van2013ontology} which used the ISO 15926 standard to model processes defined in the ISO 15288 standard. While the approach offers a general ontology for Systems Engineering, it does not cover domain specific aspects used in RV such as Mission Profile Aware Design. However, integrating this ontology into the general process flow of the presented approach may be an opportunity to explicitly describe the process using defined semantics.
In the Space Systems domain, Henning et al. use OWL ontologies to establish a system model for the integration of different modeling paradigms and model semantics \cite{henning2016spaceSystems}. The approach proposes to describe the system model and the conceptual data model using an ontology description language. Its focus is on the integration of various disciplines and their corresponding data models and does not consider MPs.
Zander et al. use DL ontologies to express features of components in the robotic domain, propagate features along compound components and to aggregate features that allow for the computation of complex features \cite{zander2015expressingAndReasoning}. Using OWL property chains for mounting point characteristics propagation as described in section \ref{sec:propagation} is inspired by this work. Also in Systems Engineering, Zhang and Luo proposed an ontology-based approach for material selection in engineering \cite{zhang2014ontologyMaterialSelections}. This approach uses a knowledge model consisting of an engineering ontology, tagged material instances and a SWRL \cite{horrocks2004swrl} rule base to encode knowledge rules for material selections. Mission Profiles are also not considered by this approach due to the different domain and application.
The application of ontologies to optimize the engineering process in collaborative development has been considered by Tudorache in \cite{tudorache2006employingOntologies}. While major contributions of this work are a framework for automatic consistency checks and the automatic propagation of changes among design models, it is not concerned with RV and therefore lacks decision support in analyses selection.

\subsection{Mission Profile Aware Design}
The general concept of \emph{Mission Profile Aware Design} (MPAD) was introduced by Jerke and Kahng in \cite{jerke2014mission} along with key differences and enhancements to existing design approaches.
Nirmaier et al. presented a methodology to extract finite state machines out of measured vehicle data and integrate them in MPs \cite{nirmaier2014mission}. This work also outlines how system robustness values can be synthesized to form component robustness values.
Katzschke et al. presented methods to propagate the constraints between design domains and introduced a cross-domain methodology for constraint transformation in \cite{katzschke2014application}.
The Reliability Knowledge Framework by NXP \cite{rongen2014reliabilityFramework} is a web based tool consisting of a Reliability Knowledge Matrix, tools and methods for Risk Assessment (RA) and Built-in Reliability (BiR) and rules and tools for applying Structural Similarity. Moreover, it contains a MP Library. It was created to connect users, reliability knowledge, data, tools and methods. Nevertheless it differs from the presented approach, as it is not ontology-based and therefore lacks the advantages of using reasoning to support the RV process. In addition to that, this approach does not provide the means to tightly integrate MP data into the RV process.

\section{Approach}
\label{sec:approach}
\begin{figure}
  \centering
  \includegraphics[width=.7\linewidth]{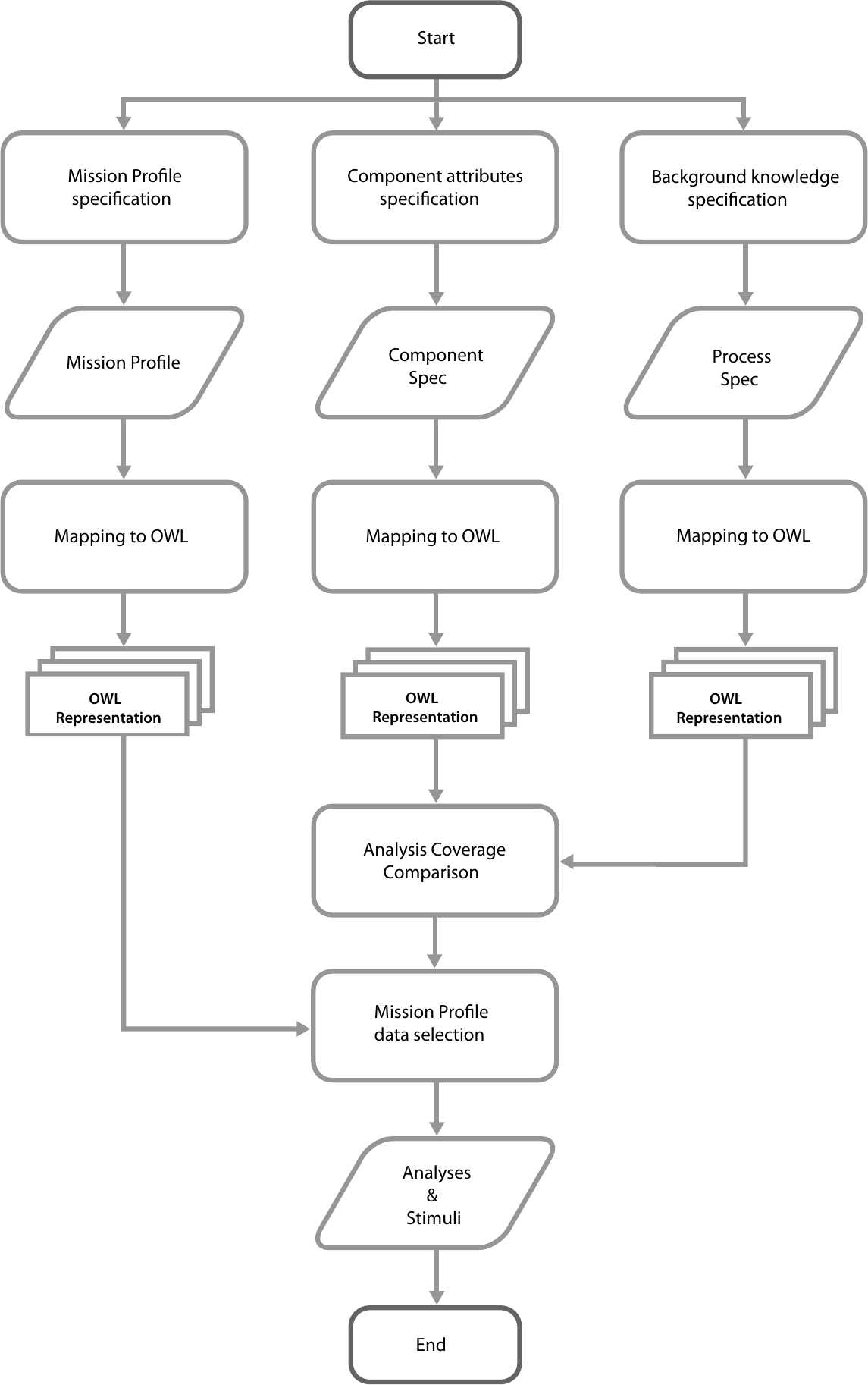}
  \caption{Overview of the approach}
  \label{fig:approach_overview}
\end{figure}
The basic idea of the approach is that developers are supported during the creation of a robustness validation plan, namely the Analysis, Development \& and Test phase (step 5 in the RV process flow \cite{byrne2008handbook}), see section \ref{sec:background_robustness_validation}. We utilize reasoning to automate the identification of required analyses and corresponding data in this phase. Properties of components and systems are formalized in OWL \cite{w3dc2012owl} and MP data is mapped to an OWL representation to allow for semantic querying for example via the SPARQL Protocol And RDF Query Language \cite{prud2008sparql}.
Furthermore, certain characteristics of components are propagated amongst neighbor components as described in the following sections.

To select an appropriate analysis, the coverage of an analysis is compared to the attributes of a component which are themselves partially propagated from mounting point characteristics. The coverage and resulting analysis to be conducted is then the foundation for the selection of the required MP data, see Fig. \ref{fig:approach_overview} for an overview of the approach.

\subsection{Mission Profile to OWL mapping}
\label{sec:owl_mapping}
To work with semantically enriched representations of MPs, a mapping mechanism is required. Apart from semantic enrichment, ontological representations of MPs are required for semantic querying. It should for example be possible to ask for specific MP data such as temperature related data. The advantage of using an ontology versus a classical approach lies in the increased maintainability, the separation of knowledge from programming and that the knowledge becomes shareable \cite{carral2013owlRules}.

A prototype of this mapping mechanism was implemented using indirect semantic programming (see \cite{paulheim2011architecture}) using the OWL API\footnote{http://owlapi.sourceforge.net/} \cite{bechhofer2003cooking} and the Scala \cite{odersky2004overview} programming language. The implementation utilizes the Eclipse Modeling Framework (EMF) \cite{steinberg2008emf} models of the MP format which were derived from the MP format XML Schema Definition (XSD) \cite{thompson2004xml} schemata. For each element in the MP a corresponding OWL individual is created and properties are added accordingly. Vectors were encoded following \cite{drummond2006sequencesInOWL}.

\subsubsection{Ontology structure}
\begin{figure}
  \centering
  \includegraphics[width=\linewidth]{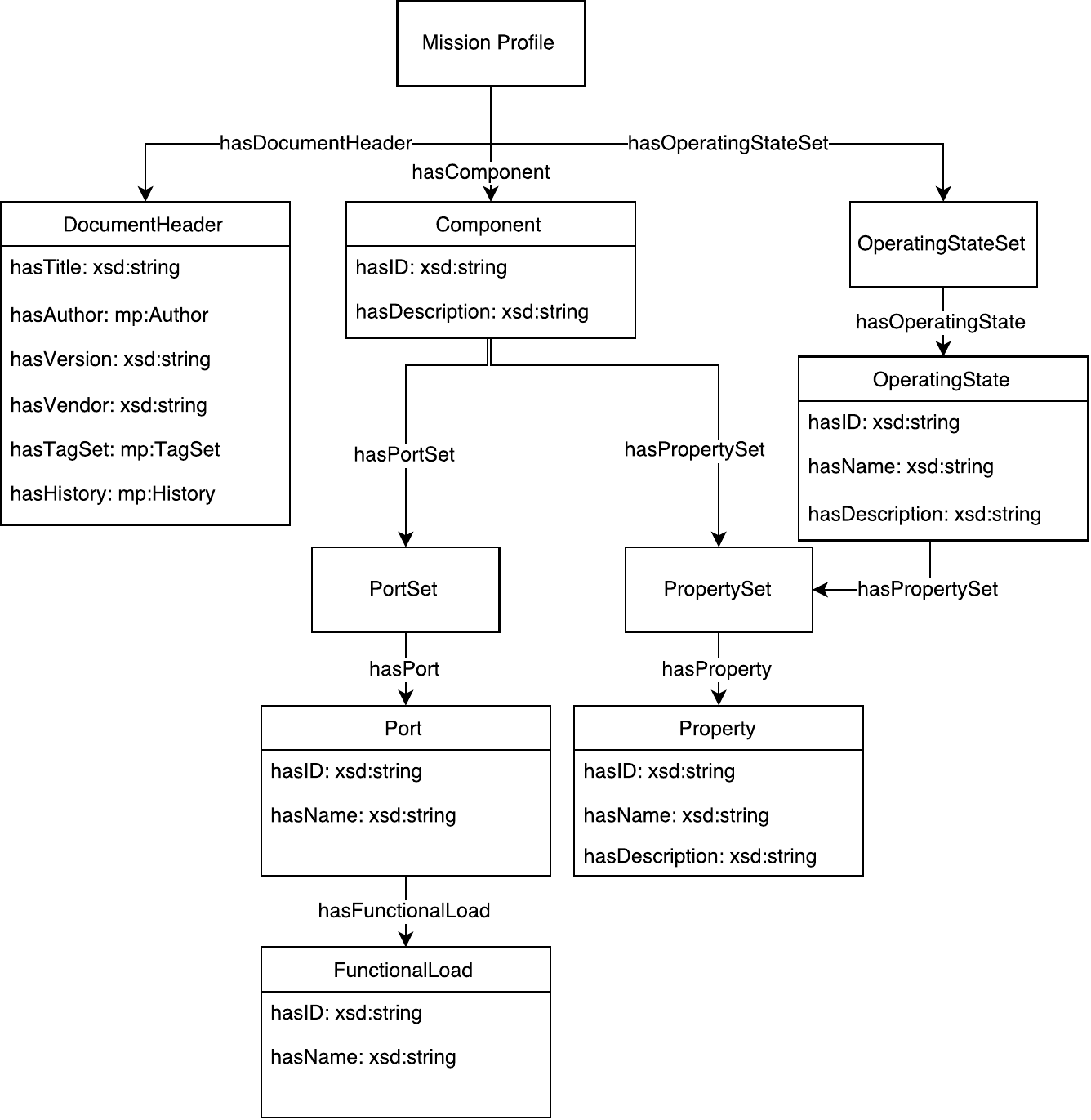}
  \caption{Excerpt of the Mission Profile ontology structure as UML class diagram}
  \label{fig:mission_profile_ontology}
\end{figure}

The MP ontology structure is based on the current MP format draft. A MP consists of three major objects, see Fig. \ref{fig:mission_profile_ontology}: (1) The \emph{DocumentHeader} (2) a \emph{Component} and (3) an \emph{OperatingStateSet}. While the DocumentHeader is used to capture meta information such as the author of the MP document or the history, the Component object is used to specify the loads for the component of a MP document. Apart from the load specifications themselves, a Component contains a \emph{PortSet} which is used to specify loads of the functional ports of a component. These ports may be electrical, mechanical, thermal or chemical connections.
Each model instance document (ABox) is connected to the MP ontology (TBox) by importing the MP ontology.
\subsubsection{Specification of functional loads}
\emph{PropertySet} and \emph{Property} objects constitute the main building block for specifying functional loads. They contain \emph{Property} objects which have an ID, a name and a description as meta information. Each property in the MP format can use a template for specifying the loads. An example would be a \emph{TemperatureTimeProfileTemplate} which is used to specify temperature profiles. Such a template is composed of a matrix which holds percent vectors to specify distributions of the temperature ranges under certain conditions and a vector to hold the actual values of the temperature ranges. These templates are mapped to their corresponding OWL representation by the mapping engine. Using templates makes the format and the MP ontology extensible by allowing to add more templates later on.
\begin{figure}
  \centering
  \includegraphics[width=\linewidth]{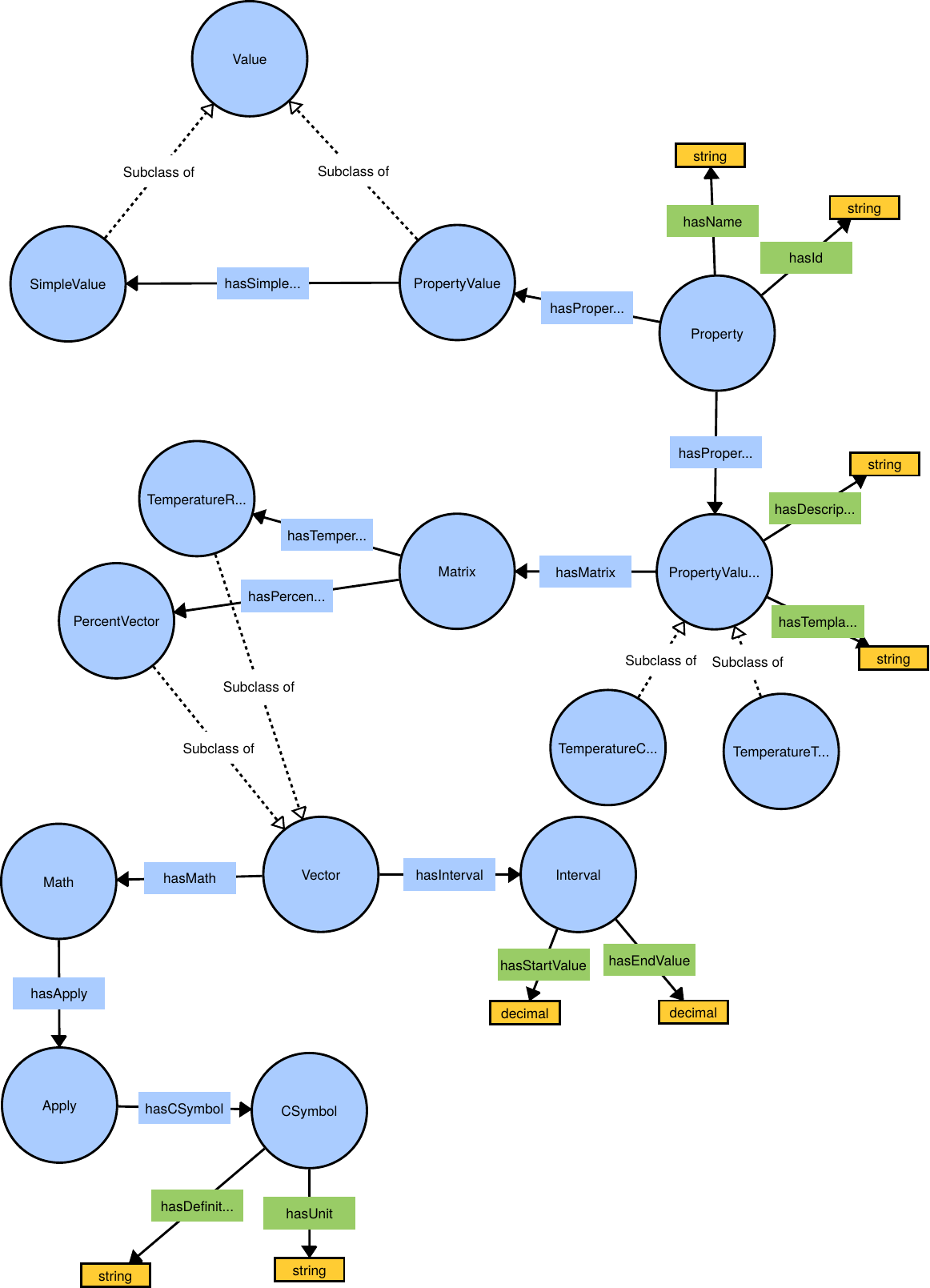}
  \caption{Model graph excerpt for property specification in VOWL}
  \label{fig:property_tree}
\end{figure}

Fig. \ref{fig:property_tree} shows some of the relationships used for property specification in the MP ontology. The shown graph is used for the specification of a functional load namely the driving profile of a business vehicle. A \emph{PropertySet} represents a set of \emph{Property} objects which themselves have a \emph{PropertyValue} object for specifying a \emph{SimpleValue} and a \emph{PropertyValueTemplate} object for specifying more complex functional loads. Each specialization of a PropertyValueTemplate is therefore a subclass of the PropertyValueTemplate concept. Of course a PropertyValueTemplate instance can be a complex structure, depending on the type of functional load that is to be specified.

\subsection{Metadata extraction}
\label{sec:metadata_extraction}
An important step in creating ontological representations of MPs is \emph{metadata extraction} (ME). ME is needed for example for the following purposes:
\begin{itemize}
\item Determine the type of data, recognize content
\item Improve search functionality
\item For provenance information
\item For visualizing data
\item Filtering data based on its content
\item For change management
\end{itemize}
An example of ME is to recognize a component's properties and associated data from MP content for classification: Suppose a MP contains a \emph{TemperatureTimeProfileTemplate} for a component. Such a template contains a matrix composed of a \emph{TemperatureRangeVector} to hold actual temperature ranges and a \emph{PercentVector} to specify the temperature distribution, Fig. \ref{fig:property_tree} shows some of the TBox information of such a template. With only the corresponding ABox facts at hand a machine does not know about the conditions in which the values were measured or even the meaning of those values at all. ME needs to be used to extract information such as the kind of vehicle that was used and other conditions. That way, a machine could decide whether a MP is within the scope of required data of a certain analysis or not. In our example, if there is a difference between a vehicle which is used for long-distance business travels and a vehicle used for family transportation ME could be used to classify the MP.

ME is required because one needs to identify the correct data for an analysis. Currently, the MP format solely provides a structure for the data but provides only little semantic information. Without this information functional load specifications for example could have incorrect units or its values might be in an invalid range. Likewise, the format in its actual state does not support to transport statements about the type of data apart from XSD templates used to specify the data itself.

Also not part of the format are statements about the context of a MP, for example the environment of a component e.g. neighbor components or the weather conditions when the data was collected. This kind of information needs to be extracted from other sources than the MP document itself.
In this work, a current draft of the MP format is used which also has some limitations. These are likely to change in the course of the autoSWIFT project.

\subsection{Analysis selection based on component properties}
\label{sec:analysis_selection}
The Handbook for Robustness Validation recommends a specific coverage for each type of analysis.
For example a \emph{Physical Stress Analysis} should be conducted for components with more than 64 pins or a length of more than 1" (2.54 cm) per side \cite{byrne2008handbook}. Such characteristics are the basis for selecting appropriate analyses in our proposed system.

\subsubsection{Modeling of components}
Foundation for identifying relevant analyses for a certain component is a component model which is provided by the developer. Such a model should define characteristics such as the pin count, self-heating capabilities and so on. See List. \ref{lst:individual} in section \ref{sec:example_application} for an example of such a component model. Ideally, the formalization of such a component model is done automatically by harvesting existing component specifications in potentially other formats than OWL.
Each component should moreover be integrated into a system model which incorporates the overall structure of the system. Very important here are the \emph{mounting points}. Mounting points are positions in a car where components can be installed. They play an important role in the propagation of mounting point characteristics which is described in section \ref{sec:propagation}.

\subsubsection{Modeling of background knowledge}
Another important step in our approach is the formalization of \emph{background knowledge}. Background knowledge is constituted of all facts that make up the guidelines of the Handbook for Robustness Validation.
For example, to express that subjects which have a pin count of more than 64 pins have a high pin count, axiom \ref{eq:PinCountValue} could be used:
\begin{equation}
\label{eq:PinCountValue}
\mathsf{\exists hasPinCountValue.(>, 64) \sqsubseteq \exists hasPinCount.HighPinCount}
\end{equation}
In case finer grained models are applied, for example when each pin is modeled as an instance of class $\mathsf{Pin}$, \emph{Cardinality constraints} \cite{rudolph2011foundations} as in axiom \ref{eq:PinCountValueObj} could be used:
\begin{equation}
\label{eq:PinCountValueObj}
\mathsf{>64.hasPin.Pin \sqsubseteq \exists hasPinCount.HighPinCount}
\end{equation}
Other background knowledge needs to be formalized likewise.

\subsubsection{Modeling of analyses coverages}
Based on the formalization of the background knowledge one can specify that components which have a high pin count have a coverage of a \emph{Physical Stress Analyis} ($\mathsf{PSACoverage}$), as expressed by axiom \ref{eq:PSACoverage}:
\begin{equation}
\label{eq:PSACoverage}
\begin{split}
\mathsf{Component \sqcap \exists hasPinCount.HighPinCount \sqsubseteq} \\
\mathsf{\exists hasCoverage.PSACoverage}
\end{split}
\end{equation}
Other coverages for other analyses need to be formalized likewise as well. According to the principle of separation of concerns, the background knowledge modeling is separated from the coverage modeling. That way, a coverage can afterwards be easily adjusted to contain specific characteristics as specified by the background knowledge modeling.

\subsection{Propagation of mounting point characteristics}
\label{sec:propagation}
Some characteristics of components are induced by other components which are locally close to the component in question. Examples for these characteristics are electromagnetic interference (EMI), temperature or vibration.\\
To model such a propagation of characteristics among neighbor components (components which are locally close to each other) complex role inclusion axioms (see \cite{kazakov2010extension}) are used, as in axiom \ref{eq:hasEMI}:
\begin{equation}
\label{eq:hasEMI}
\mathsf{mountedOn \circ closeTo \circ mountedOn^- \circ hasEMI \sqsubseteq hasEMI}
\end{equation}
By defining such an axiom, a reasoner is able to infer proper EMI values for neighbor components. An example for this might be having an electronic engine in a car which would be mounted on the engine mounting point. A component close to this mounting point will be exposed to high EMI.
\begin{figure}
  \centering
  \includegraphics[width=.8\linewidth]{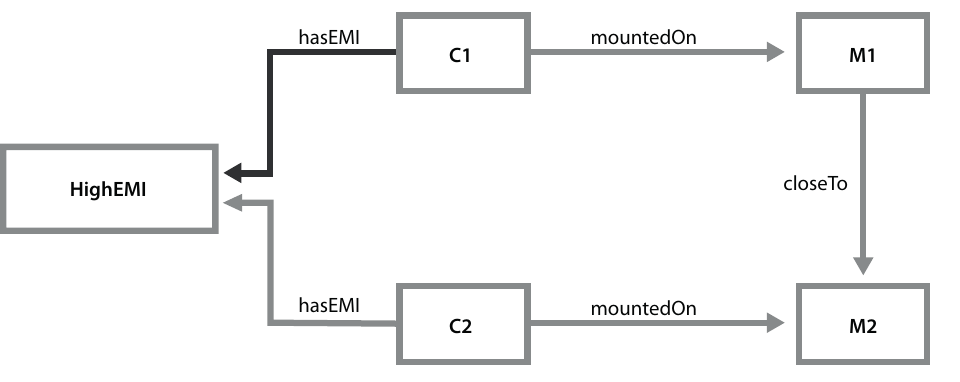}
  \caption{Example for the propagation of mounting point characteristics}
  \label{fig:propagation}
\end{figure}
Fig. \ref{fig:propagation} depicts an example where components $C1$, $C2$ are mounted on the mounting points $M1$ respectively $M2$. Component $C2$ which may for example be an electronic engine has a high EMI value. By using axiom \ref{eq:hasEMI} a reasoner can infer that $C1$ also has a high EMI value (dark grey arrow).

\subsubsection{Mounting point normalization}
For also considering manufacturer-specific mounting point definitions while at the same time maintaining the characteristics propagation we propose \emph{mounting point normalization}. The idea is for specialized installation space definitions to reference normal mounting point definitions provided by our approach. Examples for normal mounting points would be the rear of a car, the passenger compartment, the underbody, roof etc. Manufacturers could define their component mounting points and use either $\mathsf{rdfs:subClassOf}$ or $\mathsf{closeTo}$ relations to support the mounting point normalization. Of course this depends on the style of modeling. It would also be possible to realize this mechanism by creating instances of a $\mathsf{MountingPoint}$ class and then add characteristics via property specifications to those instances.

\begin{figure}
  \centering
  \includegraphics[width=.8\linewidth]{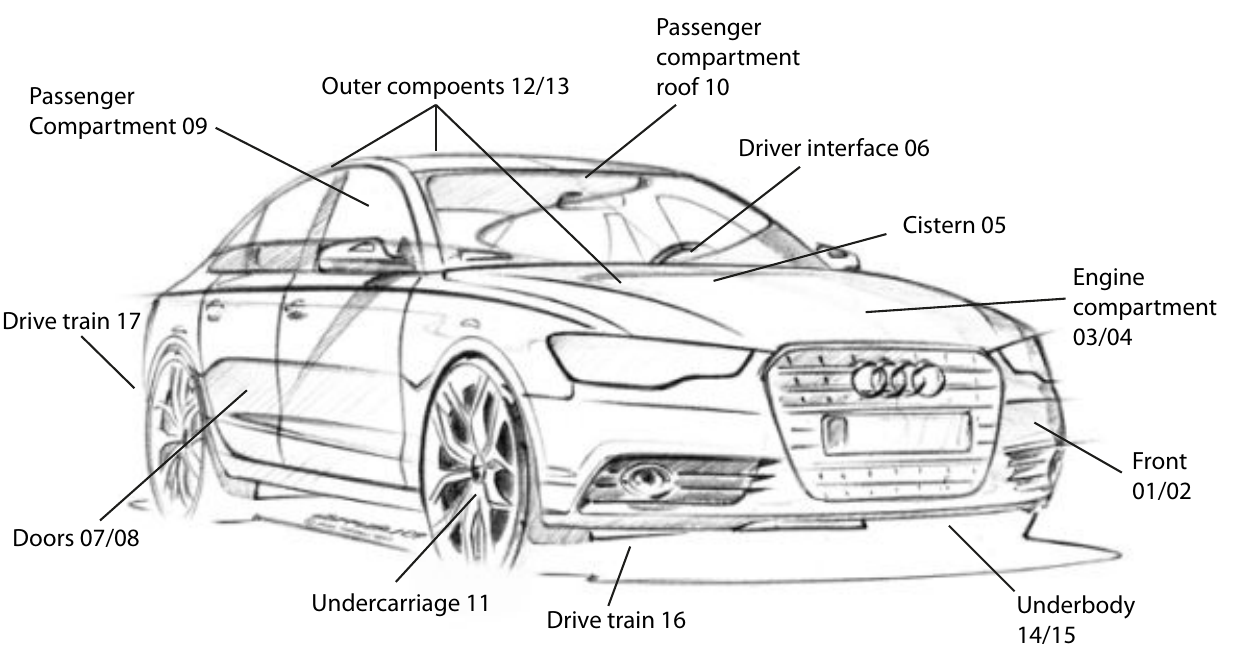}
  \caption{Mounting point definitions of an automobile, adapted from \cite{rescar2016report}}
  \label{fig:mounting_points}
\end{figure}

An example for manufacturer-specific installation space definitions would be the definitions proposed by Audi and VW based on the norm VW 80000 \cite{lv124201380000}. In the course of the ResCar 2.0 project the amount of mounting points was extend to 17, see Fig. \ref{fig:mounting_points}.
These mounting point definitions include two separate mounting points in the engine compartment (mounting point 03 respectively 04) either referencing to the whole engine compartment or to components directly mounted on the engine. These mounting points could then have the following characteristics: (1) both mounting points are $\mathsf{MountingPoint}$ instances or its instances are sub classes thereof and (2) they are $\mathsf{closeTo}$ the $\mathsf{EngineCompartment}$ mounting point. A reasoner can then infer the normalized mounting point and add corresponding characteristics to each of those manufacturer-specific mounting points.

\subsubsection{Rule-based propagation}
SWRL can be used to add rules to an OWL ontology and express certain relations which could otherwise not be expressed. Further advantages of introducing rule-based relation definitions are that rules are easier to understand because they match intuitive formulations of relations more closely. It is also easier to implement tool support for the management of the rules. Within this approach, SWRL rules are used to realize the propagation of characteristics, especially with regard to the transformation of values. A drawback of the propagation as described before in section \ref{sec:propagation} is that it is hard to derive concrete values from the materialized EMI values. Among the predicates which can be used to create a rule in SWRL is a set of so-called \emph{Built-Ins} (see \cite{horrocks2004swrl}) which can be used to implement for example comparisons, mathematical operations and string manipulations on variables.
\begin{equation}
\label{eq:rule}
\mathsf{Component(x) \land hasEMI(x,y) \land computeEMI(y, z) \rightarrow hasComputedEMI(x, z)}
\end{equation}
Equation 5 provides an example for such a rule-based value transformation where $\mathsf{computeEMI}$ in the body of the rule is a custom predicate (Built-In) which stores the result of it's computation in the unbound variable $\mathsf{z}$ which is then assigned to the $\mathsf{Component}$ via the $\mathsf{hasEMI}$ predicate in the head of the rule.
These Built-Ins can be used to model a transformation upon the propagation of values and can even be extended with custom Built-Ins implemented specifically for the computation of characteristic values. In this approach we used the SWRLAPI\footnote{https://github.com/protegeproject/swrlapi} to implement the rule-based value transformation upon propagation. This API even allows to transport custom Built-Ins which is required to create the necessary interoperability among the systems used for processing the propagation and between those systems and the systems used for the development of the actual value transformations.

\subsubsection{Rule- and threshold-based EMI check}
Using axiom \ref{eq:rule} as described in the previous section allows the computation of concrete values for checking EMI. This was an abstract description of the means used to check for EMI. We implemented a rule- and threshold-based mechanism to actually check for EMI conformance using only standard Built-Ins. The principle is to introduce \emph{comparison pairs}, assign components to them together with a maximum value for the noise signal strength. Therefore it is required to generate instances for a $\mathsf{ComparisonPair}$ class which cannot be done with OWL or SWRL alone. Once these instances are created and their properties are assigned, a rule as in axiom \ref{eq:largeRule} can be used to associate an $\mathsf{Exceeded}$ instance with a pair upon exceeding a maximum threshold.
\begin{align}
\label{eq:largeRule}
& \mathsf{ComparisonPair(p) \land} \notag \\
& \mathsf{hasFirstComponent(p, c1) \land hasSecondComponent(p, c2) \land } \notag \\
& \mathsf{hasInterferingSignal(c1, is1) \land hasInterferingSignal(c2, is2) \land } \\
& \mathsf{swrlb:add(resSig, is1, is2) \land hasMaximumInterference(p, max) \land } \notag \\
& \mathsf{swrlb:lessThanOrEqual(resSig, max) \rightarrow hasEMI(p, Exceeded)} \notag
\end{align}
Likewise, an analog rule could be defined which associates an $\mathsf{NotExceeded}$ instance to comparison pairs when a threshold is not exceeded. In that case, $\mathsf{swrlb:greaterThan}$ would be used instead of the $\mathsf{swrlb:lessThanOrEqual}$ SWRL Built-In. An opportunity to improve this check could either be for example the introduction of \emph{comparison groups} or another way to consider the locational distances of the components.

\subsection{Selecting appropriate data to support analyses}
Based on the metadata extraction described in section \ref{sec:metadata_extraction} and the analysis selection described in section \ref{sec:analysis_selection}, appropriate data can be directly passed to the analyst. This data is in form of so-called \emph{Mission Profiles} which contain all functional and environmental loads of a component, as described in section \ref{sec:mission_profiles}.

We have implemented a system prototype which automatically maps MP documents to an OWL representation to enable semantic querying of these documents, see section \ref{sec:owl_mapping}. Thereby MPs can be easily queried for specific data portions which are relevant for a certain analysis. An example might be to provide temperature and vibration data in case a \emph{Physical Stress Analysis} is conducted. The advantage of using semantic queries lies in the possibility to query for very specific data portions which are for example dependent on the environment such as the system model. Moreover, they allow to query for implicitly stated information. An example might be a query for temperature data of a certain component which is mounted on a specific location and only if the system contains a combustion engine. Following the principle of Linked Data each entity is connected by RDF links, creating a global data graph which enables the discovery of new data sources \cite{bizer2009linked}.

\section{Example Application}
\label{sec:example_application}
The Fairchild Semiconductor\textsuperscript{\textregistered} FTCO3V455A1 is a 3-Phase inverter automotive power module\footnote{https://www.fairchildsemi.com/datasheets/FT/FTCO3V455A1.pdf}. It can be used for electric power steering, electro-hydraulic power steering, in electric water pumps and electric oil pumps. To prove robustness according to the Handbook for Robustness Validation of Automotive E/E Modules, a RV plan needs to be created. Part of this plan is a set of analyses which should be conducted. To build up an ontological representation of the module, existing module models are transformed into OWL representations by a software tool. Besides harvesting the models, contextual information is ingested by the system such as the mounting point of the module and that the module is part of a mechanical system. List. \ref{lst:individual} shows a serialization of the module OWL model in Manchester syntax.

\begin{figure} 
\begin{lstlisting}[frame=single,caption=Individual in Manchester syntax,label=lst:individual,xleftmargin=3.4pt,xrightmargin=3.4pt]
Individual: FTCO3V455A1

    Types: 
        Component
    
    Facts:  
     hasPurpose  Streering,
     mountedOn  EngineCompartmentMountingPoint,
     partOf  MechanicalComponent1,
     hasHeight  29.0,
     hasPinCount  19,
     hasPowerDissipation  115.0,
     hasWidth  44.0
\end{lstlisting}
\end{figure}

Available MP data is mapped to semantically enriched representations in parallel. Metadata extraction is used on the MP data to support classification for proper data selection.
Afterwards, reasoning is used to select proper analyses and corresponding data. List. \ref{lst:inferredAxioms} shows an excerpt of the inferred axioms for the FTCO3V455A1. The EMI value is induced by another component ($\mathsf{ElectronicEngine}$) which is locally close to the FTCO3V455A1 by using the property chain described in section \ref{sec:propagation}.

\begin{figure} 
\begin{lstlisting}[frame=single,caption=Inferred axioms in Manchester syntax,label=lst:inferredAxioms,xleftmargin=3.4pt,xrightmargin=3.4pt]
Individual: FTCO3V455A1

    Types:
        Component

    Facts:
     hasCoverage  CaS_Analysis_Coverage,
     hasCoverage  EMCaSI_Analysis_Coverage,
     hasCoverage  PSA_Coverage,
     hasCoverage  E/EPaL_Analysis_Coverage,
     hasEMI  HighEMI,
     hasSiblingComponent  ElectronicEngine,
     hasSiblingComponent  FTCO3V455A1,
     needsAnalysis  CaSystems_Analysis,
     needsAnalysis  EMCaSI_Analysis,
     needsAnalysis  Physical_Stress_Analysis,
     needsAnalysis  E/E_Power_and_Load_Analysis,
     needsData  MissionProfileEMIData1,
     needsData  MissionProfilePhysicalStressData1,
     needsData  MissionProfileTemperatureData1,
     needsData  Temperature
\end{lstlisting}
\end{figure}

\begin{figure}
  \centering
  \includegraphics[width=0.7\linewidth]{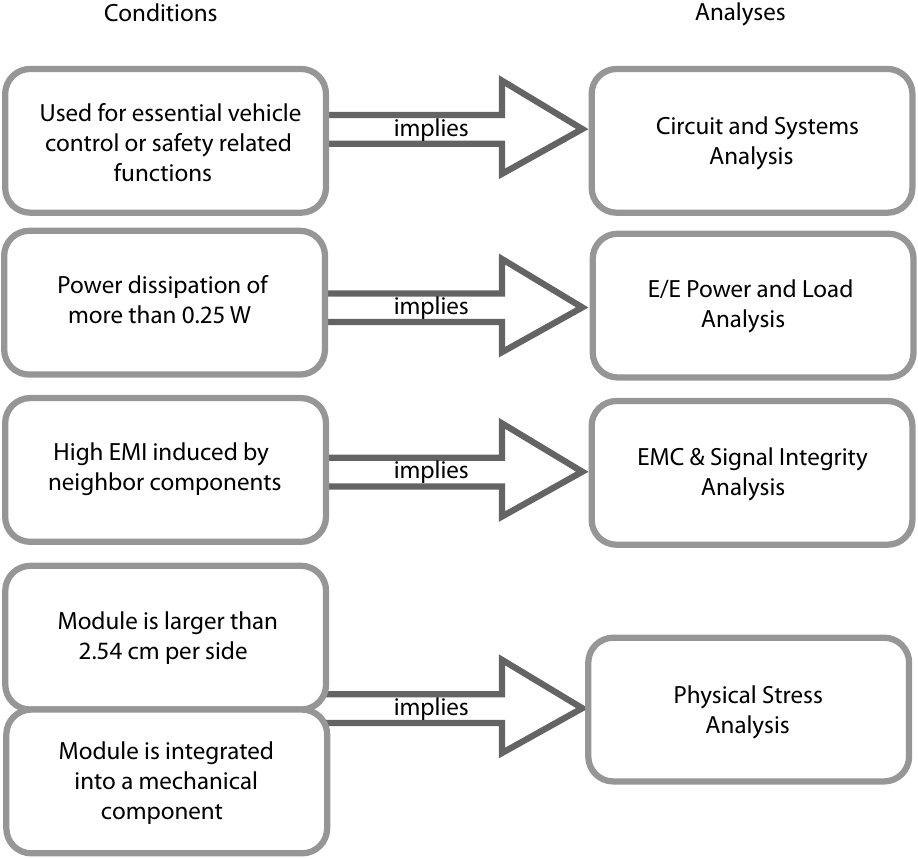}
  \caption{Analyses selection for the FTCO3V455A1}
  \label{fig:analyses_selection}
\end{figure}

Fig. \ref{fig:analyses_selection} depicts the selection of the analyses according to the approach.
The RV process is  being accelerated and errors due to manual analyses selection are avoided. Suppose that  suddenly in the development phase the specification of the inverter module slightly changes. The presented approach could then be used to easily update the RV plan accordingly. 

On a higher level, the benefit is that for a module composed of several sub modules a property chain as defined by axiom \ref{eq:needsAnalysis} can be used to aggregate the analyses which need to be conducted for the whole module.
\begin{equation}
\label{eq:needsAnalysis}
\mathsf{partOf^- \circ needsAnalysis \sqsubseteq needsAnalysis}
\end{equation}
Likewise Robustness Indicator Figures (RIFs, see \cite{byrne2008handbook}) could be aggregated to provide an overview of the overall module robustness.

\section{Generalization}
\label{sec:generalization}
\begin{figure}
  \centering
  \includegraphics[width=0.9\linewidth]{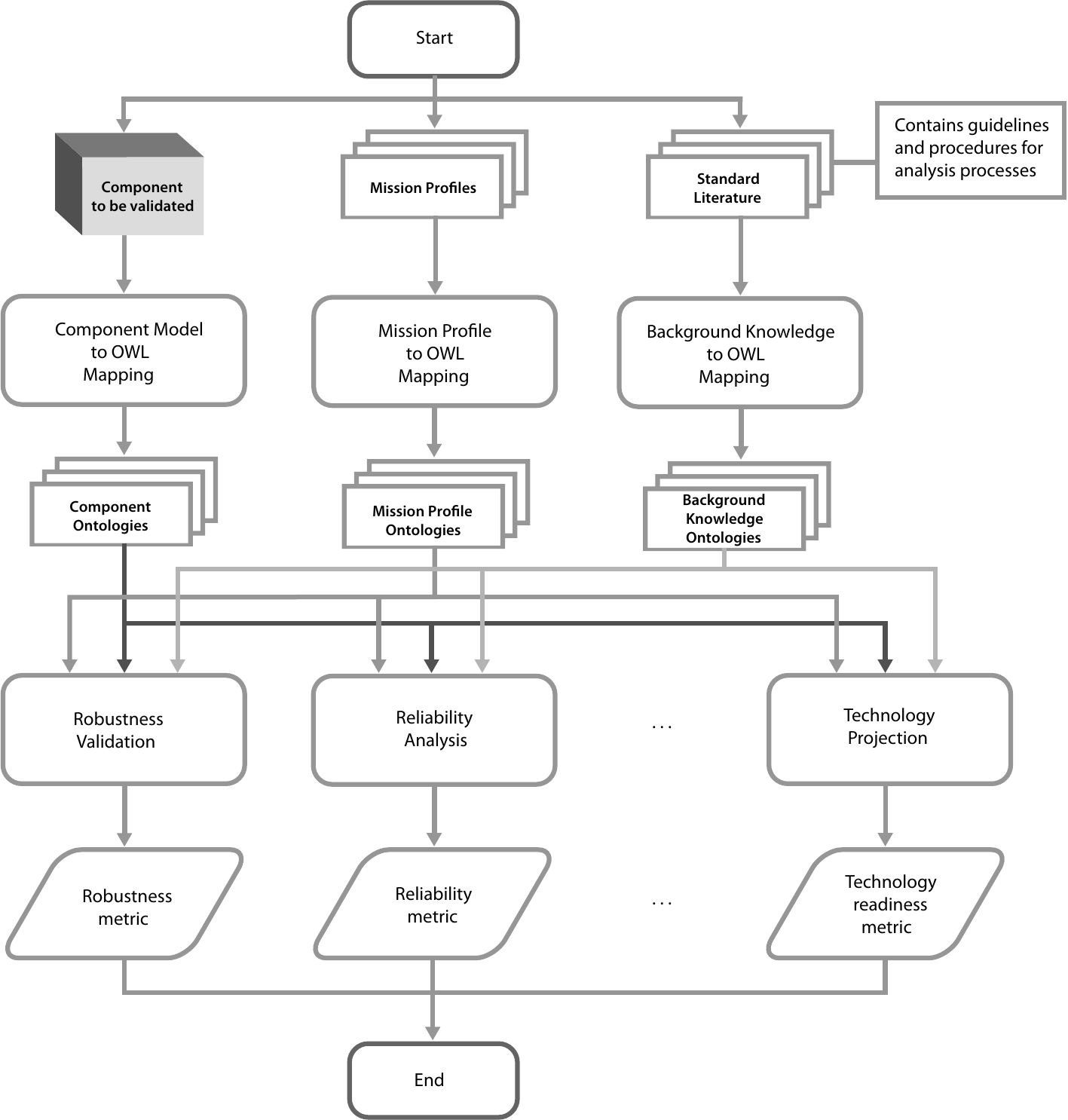}
  \caption{General process flow}
  \label{fig:generalization}
\end{figure}

It is possible to generalize the approach and apply it to other analysis methodologies, wherever analysis and corresponding data selection is done, see Fig. \ref{fig:generalization}. The foundations are a component to be examined, corresponding MP data and standard literature which contains the guidelines and procedures for the analysis process.
The core of the approach is made up by mapping engines which map component models, MPs and standard literature to their corresponding OWL representations. These representations are afterwards used as input to the analyses selection process.
Results of the various analysis methodologies are metrics while the results of the presented approach are sets of analyses and stimuli which are required for computing the metrics (see Fig. \ref{fig:approach_overview}).

\subsection{Stress test selection within AEC Q100}
The Automotive Electronics Council (AEC) released several standards which define electrical component qualification requirements, among them the \emph{AEC Q100} \cite{aecQ100}. This standard is widely used for the qualification of Integrated Circuits in the automotive domain.
The presented approach can be applied to the qualification process. One difference to the example application described before in section \ref{sec:example_application} is that stress tests are selected in opposition to analyses. Thus when applying the approach to the stress test selection these stress tests correspond to analyses having certain coverages as well which is the basis for the test selection. Another difference is, that the stress tests according to the standard define all environmental conditions required to conduct the test, so delivering MPs is not as helpful to the analyst as in analysis selection. Instead, environmental conditions for the stress tests are encoded in ontologies. Apart from that the flow is quite similar: A component model is provided as an input to the presented approach which is the basis for the stress test selection. Background knowledge is modeled in ontologies. Then, the coverages of the stress tests are compared to the the component characteristics.

\begin{table}
\centering
\label{table:stress_tests}
\begin{tabular}{|c|c|l|}
\hline
\textbf{STRESS} & \textbf{NOTES} & \multicolumn{1}{c|}{\textbf{ADDITIONAL REQUIREMENTS}} \\ \hline
\textbf{\begin{tabular}[c]{@{}c@{}}Temperature\\ Cycling\end{tabular}} & \begin{tabular}[c]{@{}c@{}}H, P, B,\\ D, G\end{tabular} & \begin{tabular}[c]{@{}l@{}}PC before TC for surface mount devices.\\    Grade 0: -55oC to +150oC for 2000 cycles or\\      equivalent.\\    Grade 1: -55$^{\circ}$C to +150$^{\circ}$C for 1000 cycles or\\      equivalent.\\       Note: -65$^{\circ}$C to 150$^{\circ}$C for 500 cycles is\\       also an allowed test condition due to legacy\\       use with no known lifetime issues.\\    Grade 2: -55$^{\circ}$C to +125$^{\circ}$C for 1000 cycles or\\      equivalent.\\    Grade 3: -55$^{\circ}$C to +125$^{\circ}$C for 500 cycles or\\      equivalent.\\ \\ TEST before and after TC at hot temperature. \\ After completion of TC, decap five devices from\\ one lot and perform WBP (test \#C2) on corner\\ bonds (2 bonds per corner)  and one mid-bond per\\ side on each device. Preferred decap procedure to\\ minimize damage and chance of false data is\\shown in Appendix 3.\end{tabular} \\ \hline
\end{tabular}
\caption*{Tab. 1: Excerpt from the AEC Q100 qualification test methods \cite{aecQ100}}
\end{table}

Tab. 1 shows an excerpt of relevant knowledge from the overview of qualification test methods. The \emph{notes} column lists some facts which can be used as criteria for the test selection based on the component model. In this example the stress test is required only for:
\begin{itemize}
\item[H] Hermetic packaged devices only
\item[P] Plastic packaged devices only
\item[B] Solder Ball Surface Mount Packaged (BGA) devices only
\item[D] Destructive test, devices are not to be reused for qualification or production
\item[G] Generic data allowed
\end{itemize}
Moreover, the \emph{additional requirements} column contains more facts which can be used as criteria as well, in this example the fact that this stress test is for surface mount devices only. The information in this column even contains knowledge about other conditions for the stress test, e.g. that the \emph{preconditioning} test should be conducted before the \emph{temperature cycling} test. This kind of knowledge is encoded in an ontology and used for the stress test selection. By using this knowledge even the order of stress tests can then be considered. In addition to the stress test selection itself, test conditions, sample sizes, number of lots and acceptance criteria can be provided to the analyst as well.

\subsubsection{Test selection in process change qualification}
The approach can also be useful when being applied to the test selection in process change qualification. Table 3 on page 18 within the AEC Q100 document \cite{aecQ100} shows which stress tests need to be conducted upon process changes. Not all tests should be considered upon a process change and not all tests need to be conducted for every device. For example upon a \emph{circuit rerouting} process change, the \emph{temperature cycle} stress test should only be conducted for peripheral routing and the \emph{power temperature cycling} test should only be conducted for devices requiring this test: "Test required only on devices with maximum rated power $\geq$ 1 watt or $\Delta T_j \geq$ 40$^{\circ}$C or devices designed to drive inductive loads.". These requirements constitute the facts for coverage comparison in the presented approach.

\section{Conclusions and future Work}
\label{sec:conclusions}
We presented an approach to tackle complexity in the RV process flow by utilizing reasoning.
Nevertheless, there are some limitations.
The Open World Assumption (OWA) imposes a limitation upon the approach: It is not possible to express that a certain component does \emph{not} have a certain property. As \cite{zander2015expressingAndReasoning} pointed out, it is possible to use \emph{closure axioms} to handle this but not without introducing further problems.
Another point is that the amount of analyses needs to be large enough for the system to be useful. If there is only a small amount of analyses, the benefit will also be small because a developer could easily create the RV plan manually.
Regarding incomplete \& incorrect knowledge, Entity Matching (EM) could be used.
The scalability of the proposed methodology depends on the efficiency of the reasoner implementations used. As the approach does not rely on any specific implementation any reasoner can be used and it might be possible to further improve the efficiency with optimizations. 

To make such a system described in this paper ready for deployment, we have identified three major necessary steps:
\begin{description}
\item[Automatic formalization of component properties] To ease the transition from the manual approach in the RV process to a more automated one, it would be beneficial to implement a system which automatically extracts component descriptions.
\item[Broadening the set of analyses] The higher the complexity of the selection process in the RV elaboration of the robustness validation plan is, the more useful will the system be. Therefore the amount of analyses which are regarded by the system should be high. It might also be an opportunity to generalize the approach to common analyses selection, see section \ref{sec:generalization}.
\item[Implementing metadata extraction] To improve the selection of appropriate data, ME for various kinds of metadata of MPs needs to be implemented.
\end{description}
By introducing an approach to automation of the RV process we were able to decrease required manual effort. The approach also takes the propagation of mounting point characteristics into account.
At the core of the presented approach we created a mapping engine for MPs to OWL representations to allow semantic querying and improve the integration of this data into the RV process. Tooling was created and will be applied within the context of a project. The resulting application framework could be used for other MPAD tasks as well.

\section*{Acknowledgement}
This article is partially supported by the BMBF project autoSWIFT (grant number 16ES0358) and the State of Baden-W\"urttemberg, Germany, Ministry of Science, Research and Arts within the cooperative graduate program EAES of the University of T\"ubingen.

%
%

\bibliographystyle{unsrtnat}
\bibliography{main}

\end{document}